\documentclass[letterpaper, 10 pt, conference]{ieeeconf}

\IEEEoverridecommandlockouts

\usepackage{epsfig} 
\usepackage{flushend}
\usepackage{amsmath} 
\usepackage{amssymb}  
\usepackage{makecell}

\title{\LARGE \bf
Automatic Detection of Myocontrol Failures\\
Based upon Situational Context Information
}

\author{Karoline Heiwolt$^{1}$, Claudio Zito$^{2}$, Markus Nowak$^{3}$, Claudio Castellini$^{3}$ and Rustam Stolkin$^{2}$
\thanks{*This work was supported by UK Engineering and Physical Sciences Research Council (EPSRC No. EP/R02572X/1).}
\thanks{$^{1}$Karoline Heiwolt is with the Computational Neuroscience and Cognitive Robotics (CNCR), University of Birmingham, UK
        {\tt\small KXH779@student.bham.ac.uk}}%
\thanks{$^{2}$Claudio Zito and Rustam Stolking are with the Extreme Robotics Lab (ERL), University of Birmingham, UK 
        {\tt\small <C.Zito,R.Stolkin>@bham.ac.uk}}%
\thanks{$^{2}$Markus Nowak and Claudio Castellini are with the Institute of Robotics and Mechatronics, DLR --- German Aerospace Center, Germany
        {\tt\small <Markus.Nowak,Claudio.Castellini>@dlr.de}}%
}

\begin{document}

\maketitle
\thispagestyle{empty}
\pagestyle{empty}


\begin{abstract}

Myoelectric control systems for 
assistive devices are still unreliable. The user's input signals can become unstable over time due to e.g. fatigue, electrode displacement, or sweat. Hence, such controllers need to be constantly updated and heavily rely on user feedback. In this paper, we present an automatic failure detection method which learns when plausible predictions become unreliable and model updates are necessary. Our key insight is to enhance the control system with a set of generative models that learn sensible behaviour for a desired task from human demonstration. We illustrate our approach on a grasping scenario in Virtual Reality, in which the user is asked to grasp a bottle on a table.
From demonstration our model learns the reach-to-grasp motion from a resting position to two grasps (power grasp and tridigital grasp) and how to predict the most adequate grasp from local context, e.g. tridigital grasp on the bottle cap or around the bottleneck. By measuring the error between new grasp attempts and the model prediction, the system can effectively detect which input commands do not reflect the user’s intention.
We evaluated our model in two cases: i) with both position and rotation information of the wrist pose, and ii) with only rotational information. Our results show that our approach detects statistically highly significant differences in error distributions with $\mathbf{p<0.001}$ between successful and failed grasp attempts in both cases.



\end{abstract}


\section{INTRODUCTION}

When designing a prosthesis, we aim to restore the full functionality of the missing body part, but as the functionality increases, it becomes more challenging to reliably control the device \cite{F,AH,AI,V,AQ}. 
The demand for advanced and dexterous control systems among patients is high. And yet, the available myocontrol options are not very popular, with the main issue being reliability \cite{F,V,W}. 

To overcome this obstacle, Gijsberts et al. \cite{C} proposed a supervised incremental learning method, called \textit{iRR-RFF}, that allows continuous adjustment of the control model with minimal training effort. iRR-RFF is effective to retain accurate control across signal shifts \cite{G,AZ}, and the computational cost of each update is not dependent on the amount of previously trained data, so that model updates remain consistently quick. This is an important efficiency benefit, that makes the method suitable for daily use. Although this mechanism stabilises reliability over time, it still relies heavily on human-driven feedback. The user will need to permanently monitor performance and intervene for updates. In order to facilitate day-to-day use of an electric prosthesis, it would be desirable to develop a complete system, that assesses its state automatically and triggers updates when needed. 
In an attempt to automate the incremental model updates, the authors in \cite{B} trained a standard linear classifier to detect failures in the sEMG input signal. The model classified examples of myocontrol use in different tasks as good versus poor control performance, and was able to match a human observer's assessments with an overall accuracy of 76.71\%.
This classification relies mostly on detecting features that are generally not desired, such as oscillatory behaviour or high accelerations. It remains unclear whether this classifier could detect a shift in sEMG patterns, that results in plausible predictions, but produces the wrong hand configuration. 

The underpinning idea of this work is that automatic failure detection should instead be able to spot every instance, where the myocontrol output does not match the user's intention. Our hypothesis is that we can detect more subtle shifts in the control mapping by incorporating \textit{situational context information}. In the reach-to-grasp scenario, the hand configuration we choose greatly depends on the shape of the object and how we position our hand relative to the object (i.e. the local context). From human demonstration we learn the relation between context and reach-to-grasp trajectories as a set of generative models. These models are then used as a prior estimate of the user intention in similar local contexts. By comparing new motions to our models' predictions, we can assess which of the movement commands do not reflect the intention and are more likely caused by a myocontrol failure.
We developed and demonstrated our approach in a virtual reality (VR) simulation, where both patients and able-bodied users can control a model of a prosthetic hand. The simulation provides a highly controlled test environment, and allows us to accurately measure, record, evaluate and visualise the user's movements in the virtual scene. 

This paper is structured as follows. We first describe the implementation of the VR simulation. We then show how to construct the proposed approach for learning a situational context model. Finally, we demonstrate that our approach appropriately distinguishes between sets of successful and failed control performance, and can be used to detect myocontrol failures.

\section{Method}
 
\subsection{VR Simulation}\label{VRSection}

\subsubsection{Materials}

The virtual environment was developed using the Unity 3D game engine and was displayed on the HTC Vive Virtual Reality Headset. One HTC Vive Tracker was used for positional tracking of the user's right forearm. The sEMG data was gathered using the Myoband by Thalmic Labs, a stretchable bracelet fitted with 8 EMG sensors.

\subsubsection{Myocontrol Implementation}

We specifically simulate control of transradial prostheses (i.e. below the elbow joint). 
The virtual prosthesis is modelled as a biological hand with 20 degrees of freedom (DoF), that are grouped into two degrees of control (DoC) for the purpose of this work, namely coupled flexion of the thumb, index finger and middle finger, and coupled flexion of the ring finger and little finger.
Using the iRR-RFF as in \cite{C}, a model is trained to predict two normalised values, representing the proportional activation of each DoC, from 8-dimensional sEMG patterns. This activation vector is predicted at a frequency of 200 Hz. 
At every rendered frame of the simulation, we translate the latest available activation vector into 20 joint angles using a mapping matrix. 
The baseline hand configuration for a zero activation vector corresponds to the resting position (configuration \text{a)} in Fig. \ref{Grasps}).
\subsubsection{Object Interaction}
\begin{figure}[!t]
\centering
\vspace{0.2cm}
\includegraphics[width=1.5in]{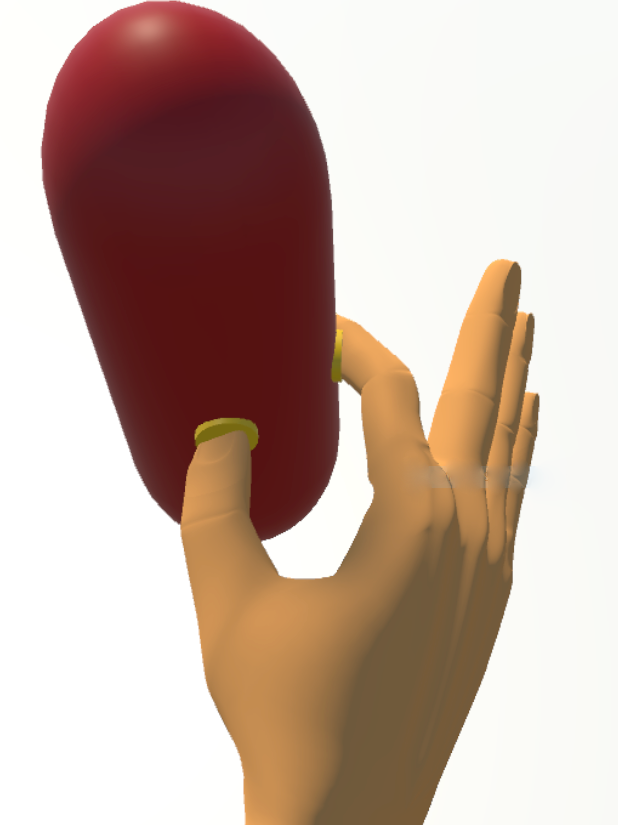}
\caption{Example of a stable grasp according to our custom
grasp stability rules in the VR simulation. Two contact points are marked in yellow on the object surface. The angle between their contact normals is 90 degrees.}
\label{Stable}
\end{figure}

Physics engines are unstable when multiple contacts act on an object, such as in grasping (e.g. \cite{BP}).
We thus rely on the physics engine only for collision detection, but allow penetration
and specify custom rules for grasp stability. 
We classify a grasp as stable if there are at least two contacts with a minimum angle of 90 degrees between their normals (see Fig. \ref{Stable}). The object is then simply attached to the wrist's coordinate frame and moves with it, as long as the condition is satisfied. This criterion was experimentally tuned to appear as natural as possible and only fulfills the purpose of enabling basic interaction with a rigid object. 

\subsection{Failure Detection}
\label{CostFunctionSection}
We aim to detect potential failures in the myocontrol by learning a set of generative models that encodes a correct behaviour for attempting a grasp. Our training data is represented as a set of $N$ states recorded along all the demonstrated grasping trajectories:
\begin{eqnarray}
\label{SetDefinition}
D = \left \{  s_{j} : s_{j} \in \mathbb{R}^{3} \times \mathbb{H} \times \left [ 0,1 \right ]^{k} \right \}_{j \in \left [ 1,N \right ]}\, ,
\end{eqnarray}
where each state $s_{j}$ consists of the position $p_{j}$ $\in \mathbb{R}^3$ of the wrist in space, the quaternion $q_{j}$  $\in \mathbb{H}$ representing the orientation of the wrist, and a vector $\mathbf{M}_{j}$ $\in \left[ 0,1 \right ]^k $ of activation values for $k$ degrees of control that encode the hand configuration.

The quality of our predictor is evaluated as a normalised weighted mean squared error (MSE) between a new state $x$ and our model's prediction based on the training data $D$. The normalised weighted MSE can be seen as the expected value of the squared error function of our model. We thus define an error function $C$ as follows:  
\begin{eqnarray}
C\left ( x, D\right ) = \frac{1} {\sum_{j = 1}^{N} w{_{j}}} \,\, \sum_{j = 1}^{N} \text{MSE}\left ( \mathbf{M}_{x}, \mathbf{M}_{j} \right )  w_{j}\,.
\label{CostEquation}
\end{eqnarray}
The difference in hand configurations is measured by the \text{MSE} between two activation vectors:
\begin{eqnarray}
\text{MSE}\left ( \mathbf{M}_{x}, \mathbf{M}_{j} \right) = 
\frac {\langle \left( \mathbf{M}_{x} - \mathbf{M}_{j} \right), \left( \mathbf{M}_{x} - \mathbf{M}_{j} \right) \rangle }  {k} \, .
\end{eqnarray}
The angled brackets $\langle \, , \rangle$ are used to denote the inner product between two vectors.
That way, the MSE compares activations for each DoC and appropriately summarises the total degree of divergence in a single value between $0$ and $1$, where $0$ expresses perfect congruence.

This error is calculated between $x$ and each state $s_{j}$ in $D$. But for the overall error of $x$, we want to compare $x$ only to previously demonstrated states in a similar context, that is states in which the prosthesis was positioned relative to the object in a similar way. Therefore we attach a weight $w_{j}$ to each training state, that represents its relative importance with respect to $x$. This weight is a function of both the spatial and angular distance between that training state and the current state $x$:
\begin{subequations}
\begin{eqnarray}
w_{j} = w_{\text{position}}\left ( p_{x}, p_{j}\right ) \times 
w_{\text{rotation}}\left ( q_{x}, q_{j}\right )
\end{eqnarray}
\text{\newline with}
\begin{eqnarray}
 w_{\text{position}}\left ( p_{x}, p_{j}\right) =
\begin{cases}
   e^{-\alpha \,\, d \left(p_{x}, p_{j}\right)^2} & \text{if } d \left(p_{x}, p_{j} \right)\leq \delta\\
   0 & \text{otherwise}
\end{cases} \\
w_{\text{rotation}}\left ( q_{x}, q_{j}\right) =
\begin{cases}
   e^{-\beta \,\, \theta \left(q_{x}, q_{j}\right)^2} & \text{if } \theta \left(q_{x}, q_{j}\right)\leq \phi\\
   0 & \text{otherwise}
\end{cases}
\end{eqnarray}
\text{\newline where  $d \left(p_{x}, p_{j}\right)$ is calculated as the euclidean distance be-}
\text{\newline tween the two points, and}
\begin{eqnarray}
 \theta \left(q_{x}, q_{j}\right)= \cos^{-1} \left( \frac{ \langle  q_{x}, q_{j} \rangle}{\sqrt{\left \langle q_{x},q_{x} \right \rangle} \times \sqrt{\left \langle q_{j},q_{j} \right \rangle}} \right ).
\end{eqnarray}
\end{subequations}

\begin{figure}[!t]
\centering
\vspace{0.2cm}
\includegraphics[width=1.8in]{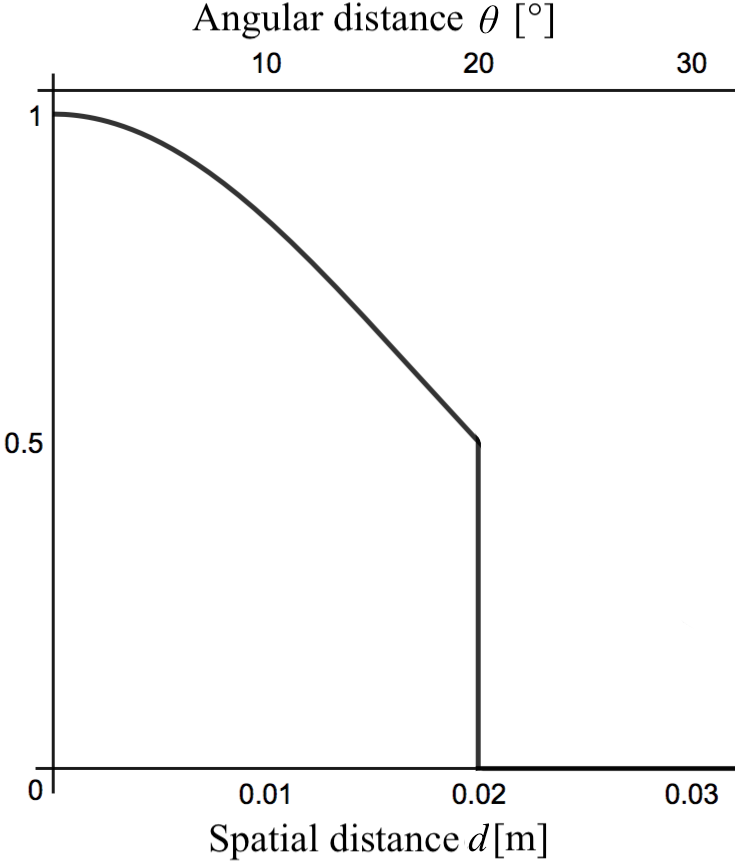}
\put(-150,28){\rotatebox{90}{$w_{\text{position}}(d)$ / $w_{\text{rotation}}(\theta)$}}
\caption{Weight functions $w_{\text{position}}$ with respect to the lower x-axis and $w_{\text{rotation}}$ with respect to the upper x-axis, plotted for the parameter values $r = 0.25$, $\delta = 0.02$m, $\phi = 20^{\circ}$, $\alpha = 1733$ and $\beta = 0.001733$.
Both functions implement the same decrease of relative importance for points within their respective cut-off distances $\delta$ and $\phi$. In this example the values range between $0.5$ and $1$, so that the combined weight $w_{j}$ can assume values between $0.25$ and $1$.}
\label{WeightPlot}
\end{figure}

These two weighting terms decrease exponentially with the square distance in space and in angle respectively. Additionally, training states further than the cut-off distances $\delta$ and $\phi$ from $x$ are excluded entirely. Thus, these two parameters define a focal area (or area of similar context) around $x$, in which training states are considered to be relevant.
Parameters $\alpha$ and $\beta$ can be tuned to specify how much points on the edge of the focal area contribute to the evaluation relative to points in the centre:
\begin{subequations}
\begin{eqnarray}
\alpha = -\frac{\text{ln}(\sqrt{r}) }{(\delta)^2} \\
\beta = -\frac{\text{ln}(\sqrt{r}) }{(\phi)^2}
\end{eqnarray}
\end{subequations}
where $r$ is the relative contribution ratio or relative importance of training states on the edge of the focal area (for example see Fig. \ref{WeightPlot}).

Then, we integrate all those single state comparisons into one overall error function $C$ for the state $x$. In Eq.~(\ref{CostEquation}) this is computed as the normalised weighted mean of the differences in hand configurations, weighted by their similarity in context, rather than a sum or an arithmetic mean over the comparisons.
The weighted mean appropriately expresses the overall divergence of the hand configuration from previous demonstrations in a similar context. It is not distorted by the amount of available data and takes into account the relative importance of training states.


In summary, the error function given in Eq.~(\ref{CostEquation}) can be used to calculate a mismatch between the user's motion and the model's prediction in a similar context. The learned models can be interpreted as an estimate of user intention, thus the error values also encode how much the performed action differs from the command we assume the user might have attempted. We can also infer whether the activation vector in the new state is likely an accurate reflection of the user’s intention (low values) or rather a failure in the myocontrol (high values). To do so, a threshold can be defined, above which states are classified as myocontrol errors.

\subsection{Experimental Method}

The ability of the proposed function in Eq.~(\ref{CostEquation}) to detect failures was analysed on a reach-to-grasp task. One set of training data and two sets of test data (a success condition and a failure condition) were collected in a single session by the same able-bodied subject, using the VR simulation described in section \ref{VRSection}.

\subsubsection{Procedure}

\begin{figure}[!t]
\centering
\vspace{0.2cm}
\includegraphics[width=1.5in]{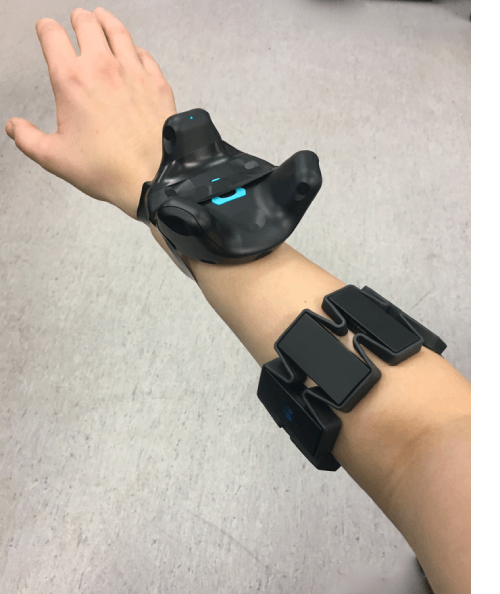}
\put(-50,100){a)}
\put(-15,65){b)}
\sbox0{\footnotemark}
\caption{Placement of the sensors on the user's arm: a) The Vive Tracker is placed just below the wrist joint and secured using adhesive tape, b) The Myoband is placed approximately 5cm below the elbow joint, where it picks up electrical signals at 8 positions around the forearm. Exact placement of the electrodes is not necessary for regression model approaches, as long as they cover different muscles around the forearm (e.g. \cite{C,AK,D}).}
\label{Setup}
\end{figure}

The sEMG bracelet and the Vive tracker were placed on the user's forearm as shown in Fig. \ref{Setup}.

\begin{figure}[!t]
\centering
\vspace{0.2cm}
\includegraphics[width=2.2in]{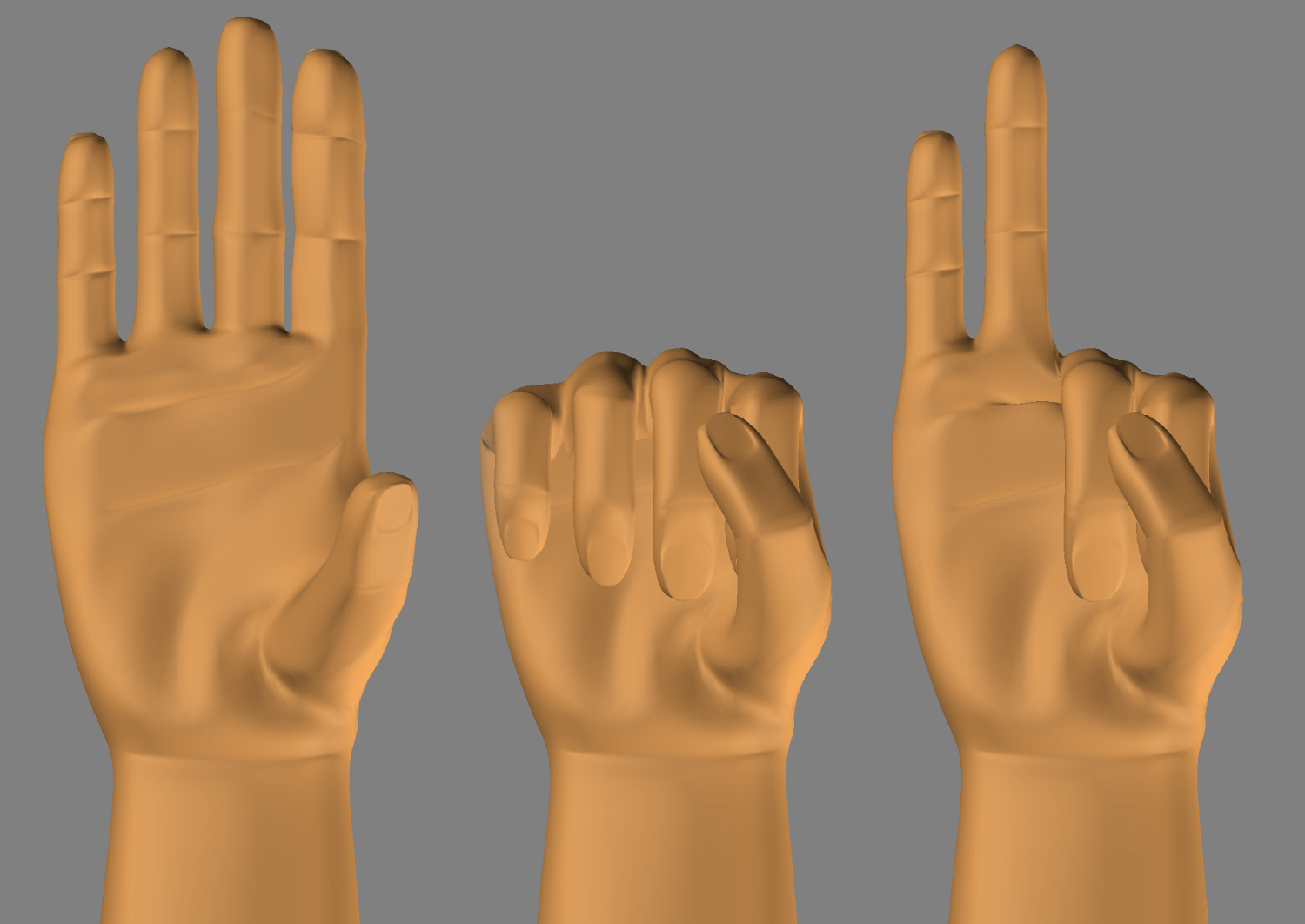}
\put(-135,8){a)}
\put(-80,8){b)}
\put(-30,8){c)}
\caption{Hand configurations used for training in the experiment: \newline a) Resting position b) Power grasp c) Tridigital grasp.}
\label{Grasps}
\end{figure}

A myocontrol model was trained on the three different hand configurations shown in Fig. \ref{Grasps}. Just as in \cite{C} and \cite{D}, the subject performed the desired grasps by copying the visual cues in Fig. \ref{Grasps} at the maximal comfortable level of force. The resulting sEMG activation pattern was labeled with the two-dimensional activation array corresponding to the displayed grasp. 
In order to achieve an accurate reflection of the user's intention, the model was carefully adapted incrementally until the subject was satisfied with the control performance and subjectively perceived no control failures. In total, each grasp was recorded eight times, four times in a relaxed position with the right arm held close to the body and bent by 90 degrees at the elbow, and four times with the arm extended forward and the palm facing inward. Once the system was trained, the subject put on the head-mounted display and started the data collection. 

For the training data, the subject performed a total of 22 trials of the grasping task. In each trial a capsule-shaped object appeared on a table and within reaching distance of the user in the virtual scene. The subject then reached for the object and picked it up, while the application recorded the trajectory and configuration of the virtual hand continuously until the first stable grasp on the object was achieved. The capsule shape was chosen, as it resembles many real life demands (such as picking up a bottle), which can be solved in a number of different ways, including at least two different grasp types (power grasp and tridigital grasp) and several different contact regions and arm configurations. So instead of repeatedly performing the same motion, the subject was encouraged to demonstrate a variety of different grasp solutions on the object. For a visualisation of the training data set, see the left panel of Fig. \ref{DemoTrajectories}.

Because this training data serves as a sample of natural grasps a human user would attempt, it is essential to avoid accidentally recording any myocontrol failures within this sample. Thus, all indistinct trials, that is trials in which the myocontrol did not exactly reproduce the subject’s intention, were discarded immediately, per the judgment of the subject. Twice during this data collection a single update was incrementally added to the myocontrol model to counteract performance degradation.

Next, we collected test data for the validation of the proposed error function.
The success condition is a sample of successfully executed grasps, representing the desired state of a reliable myocontrol system, and containing no myocontrol failures. For this sample 8 trials were collected in exactly the same way as described above, where the execution accurately reflects the subject's intention. For the failure condition, we aimed to collect data, that represents the myocontrol failures in the unreliable state of the myocontrol system following a shift in the sEMG inputs. A second myocontrol model was trained exactly as before, but this time the labels (i.e. the activation vectors) for the grasps b) and c) in Fig. \ref{Grasps} were swapped. By intentionally mis-labelling the grasps, we effectively trained an action to be associated with a different intention. The resulting activations and model predictions were plausible, but produced a different hand configuration from what the user intended. Another 20 trials were recorded with this second myocontrol model.

\subsubsection{Data Analysis}

\begin{figure}[!t]
    \centering
    \vspace{0.2cm}
    \includegraphics[width = .45\linewidth]{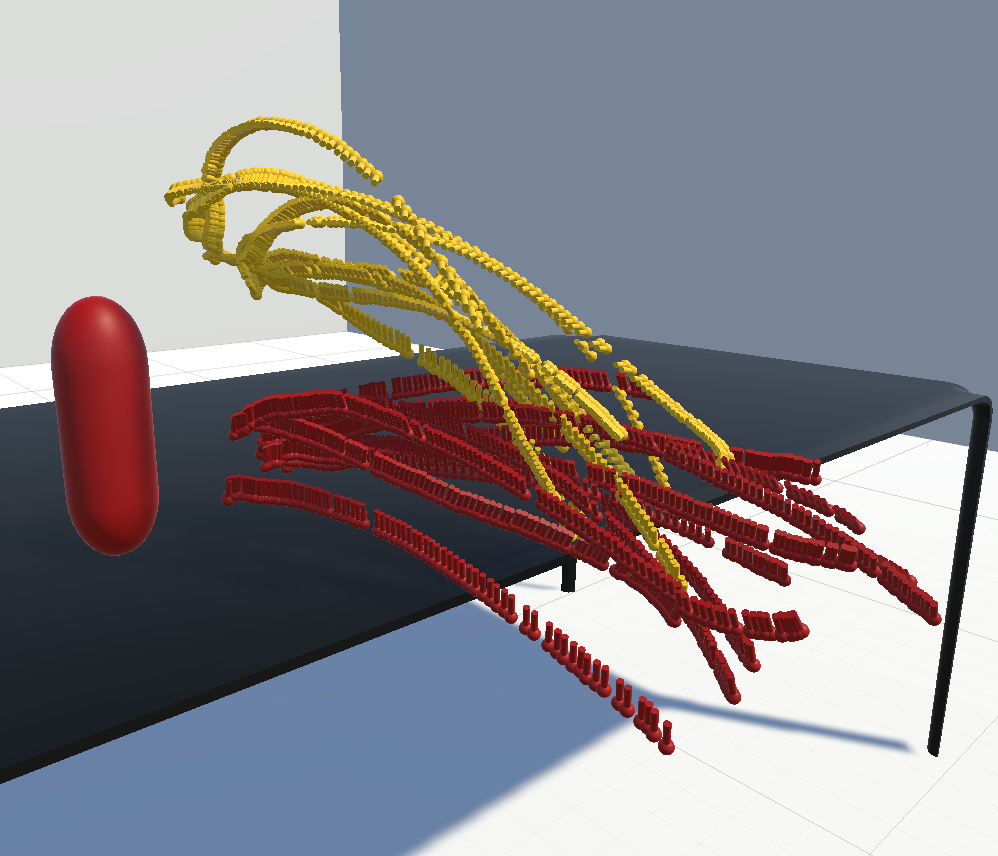}
    \includegraphics[width = .45\linewidth]{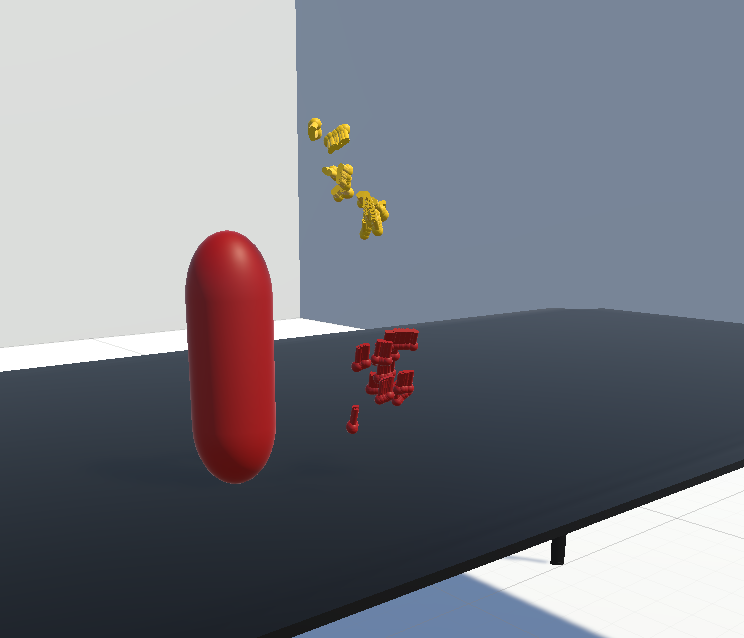}
    \caption{Visualisation of the training data. The left panel shows 2789 states sampled along the whole length of all demonstrated trajectories. The right panel shows the 440 target area training states, representing only the last 200ms of each trajectory. Trajectories pictured in yellow ended in a tridigital grasp, red trajectories ended in a power grasp.}
    \label{DemoTrajectories}
\end{figure}

The three data samples were represented as arrays of states (as defined in Eq. (\ref{SetDefinition})), sampled at approximately every 12 milliseconds along each trajectory. We then conducted two evaluations of both experimental conditions using the  function specified in Eq.~(\ref{CostEquation}).
For the first evaluation, we calculated the value of the function for every single test state in both experimental conditions based on the full training data; a total of 2789 training states, shown in the left panel of Fig \ref{DemoTrajectories}. 

The second time, we calculated the errors based on a subset of the training data, containing only the last 20 recorded data points before the stable grasp was reached in each demonstrated trial. This smaller \textit{target area training set} contains a total of 440 training states representing roughly the last 200ms of each trajectory, as shown in the right panel of Fig \ref{DemoTrajectories}.
This second evaluation was conducted, because in qualitative observations, most trajectories in both conditions started in a very similar way and only diverged into different hand configurations close to the object. We expected that the difference in error distributions between the conditions also becomes more pronounced closer to the object surface. Thus, evaluating the test data only on the smaller but more determining and relevant target area training sample should suffice to distinguish between the two conditions. Finally, all test states for which there were not enough training states $n$ available in a similar context for an evaluation ($n<5$) were excluded from further analyses. All function parameters were tuned experimentally and kept fixed throughout these experiments (Table \ref{Parameters}). For an overview of the final sample sizes, see Table \ref{DataOverview}.

\begin{table}[!t]
\caption{Parameter Choices for the Error Function}
\label{Parameters}
\centering
\begin{tabular}{c|c|c|c|c|c|c}
$k$ & $n$ & $r$ & $\delta$ & $\phi$ &$\alpha$ & $\beta$\\
\hline
$2$ & $5$ & $0.25$ & $0.02$m & $20^{\circ}$ & $1733$ & $0.001733$\\
\end{tabular}
\end{table}

\begin{table}[!t]
  \centering
  \caption{Overview of test sample sizes based on both positional and orientational information of the wrist.}
  \label{DataOverview}
    \begin{tabular}{l c c}
    & \makecell[c]{Success\\condition} & \makecell[c]{Failure\\condition}\\ \hline
    Number of grasp trials &  8 & 20 \\
    \hline
    Total number of sampled test states & 1099 & 2066 \\
    \hline\makecell[l]{Number of evaluable test states based\\on the full training data} & 605 & 765 \\
    \hline
    \makecell[l]{Number of evaluable test states based \\ on the target area training data} & 306 & 338
    \end{tabular}
\end{table}

We then compared the sets of calculated error values across the two conditions. Overall differences in the distribution of errors were analysed using statistical tests. Furthermore, we closely examined the absolute range of the values, as this feature in particular determines how appropriate a threshold would be as a final failure detection criterion. 
One possible concern of this approach is, that single trials could be over-represented within their samples. Because of different movement speeds during the data collection, the number of states within single trials varies, and we need to ensure that the overall results in the direct comparison are not distorted, but that similar results are obtained when the data is grouped into trials for analysis. Therefore, we also examined the overall distributions, the mean error and the ranges along each trial.
It is important to note however, that the error  is intended to be interpreted with respect to the single current state, and this grouped analysis only serves as a validation of direct comparison results.
And finally, we tuned an exemplary threshold value to the results and determined the sensitivity and specificity of the eventual failure classification of our test data set.

\begin{figure*}[!t]
    \centering
    \vspace{0.2cm}
    \includegraphics[width = 6.5in]{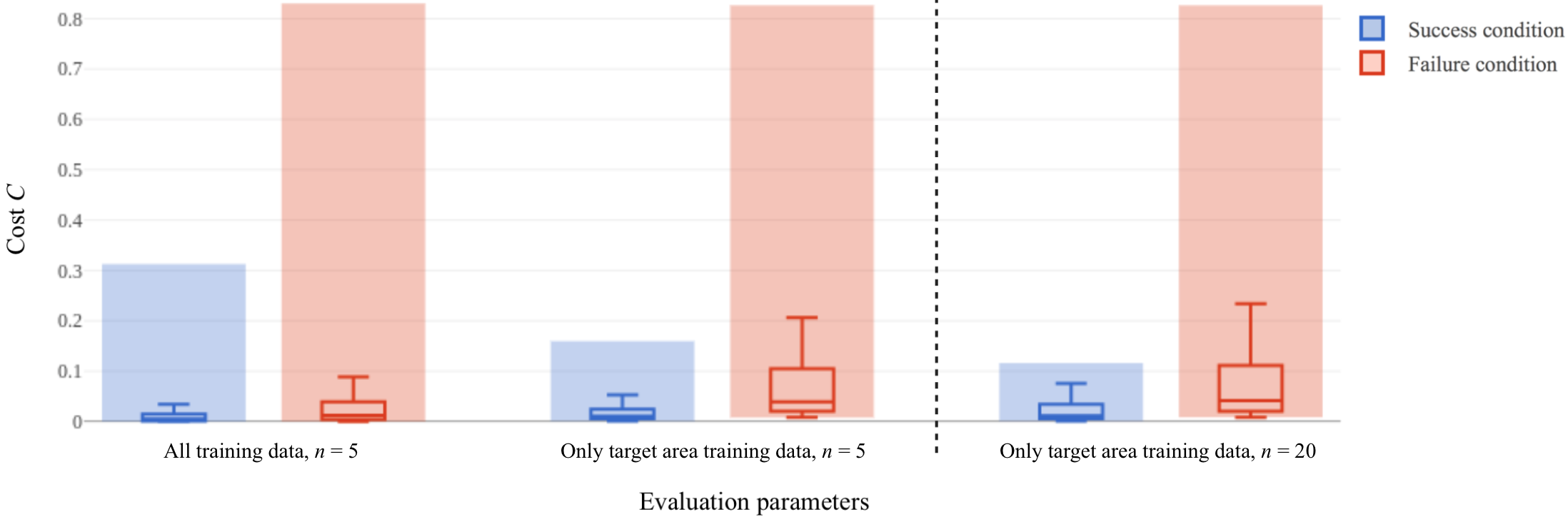}
    \caption{Comparison of error distributions between the two experimental conditions when both positional and rotational information were available. The chart shows Tukey boxplots, where the whiskers represent data within 1.5 inter-quartile-ranges of the first and third quartile. The bars illustrate the absolute range of error values. The two groups to the left display the results obtained in evaluation 1 and 2, while the the group on the right represents just one example of how error distributions can be influenced by tuning other variables of the error function.}
    \label{Boxplot}
\end{figure*}

To translate this approach into a mobile prosthesis, the device needs to be fitted with a depth camera and on-board positional and rotational tracking that does not require external reference (for example inertial tracking, e.g. \cite{BY}). Since this positional information is much harder to obtain than orientational information, we also assessed the distribution of errors and the potential failure detection performance based only on the orientation of the wrist, in case there is no positional information of the wrist available at all.

\section{Results}

\subsection{Based on both positional and rotational information}

The distributions of the errors for both conditions in both evaluations are shown in the left two groups of Fig. \ref{Boxplot}. Most of the data accumulates around the low end of the spectrum. Overall, however, the failure condition yielded higher values than the success condition (Table \ref{ResultsTable}). This difference was further analysed using a permutation test. 
\begin{table}[!t]
    \centering
    \vspace{0.2cm}
    \caption{Comparison of error distributions based on both positional and orientational information of the wrist.}
    \label{ResultsTable}
    \begin{tabular}{r c c}
    \textbf{Evaluation 1} & Success condition & Failure condition \\
    \hline
    $M$ &  0.0129 & 0.0437 \\
    $SD$ & 0.0269 & 0.1066 \\
    Median & 0.0054 & 0.0120 \\
    \hline
    $p$ & \multicolumn{2}{c}{$<$0.001} \\
    \end{tabular} \\
    \vspace{0.3cm}
    \begin{tabular}{r c c}
    \textbf{Evaluation 2} & Success condition & Failure condition \\
    \hline
    $M$ & 0.0247 & 0.0948 \\
    $SD$ & 0.0317 & 0.1489 \\
    Median & 0.0104 & 0.0391 \\
    \hline
    $p$ & \multicolumn{2}{c}{$<$ 0.001}
    \end{tabular} \\
    \vspace{0.3cm}
    \raggedright{\textit{Note.} Values reported in this table are the mean (\textit{M}), standard deviation (\textit{SD}), and the median of all calculated errors in each experimental condition, as well as the one-sided \textit{p}-value of a permutation test between the conditions after 5000 permutations.}
\end{table}
Specifically, we used a MatLab implementation by Ehninger \cite{BU}, based on the corrections proposed by Phipson and Smyth \cite{BR}. The test detected statistically highly significant differences in error distributions with \textit{p} $<$ 0.001 after 5000 permutations for both evaluations.

\begin{figure}[!t]
    \centering
    \vspace*{-0.3cm}
    \hspace{-0.7cm}
    \includegraphics[width = 2.8in]{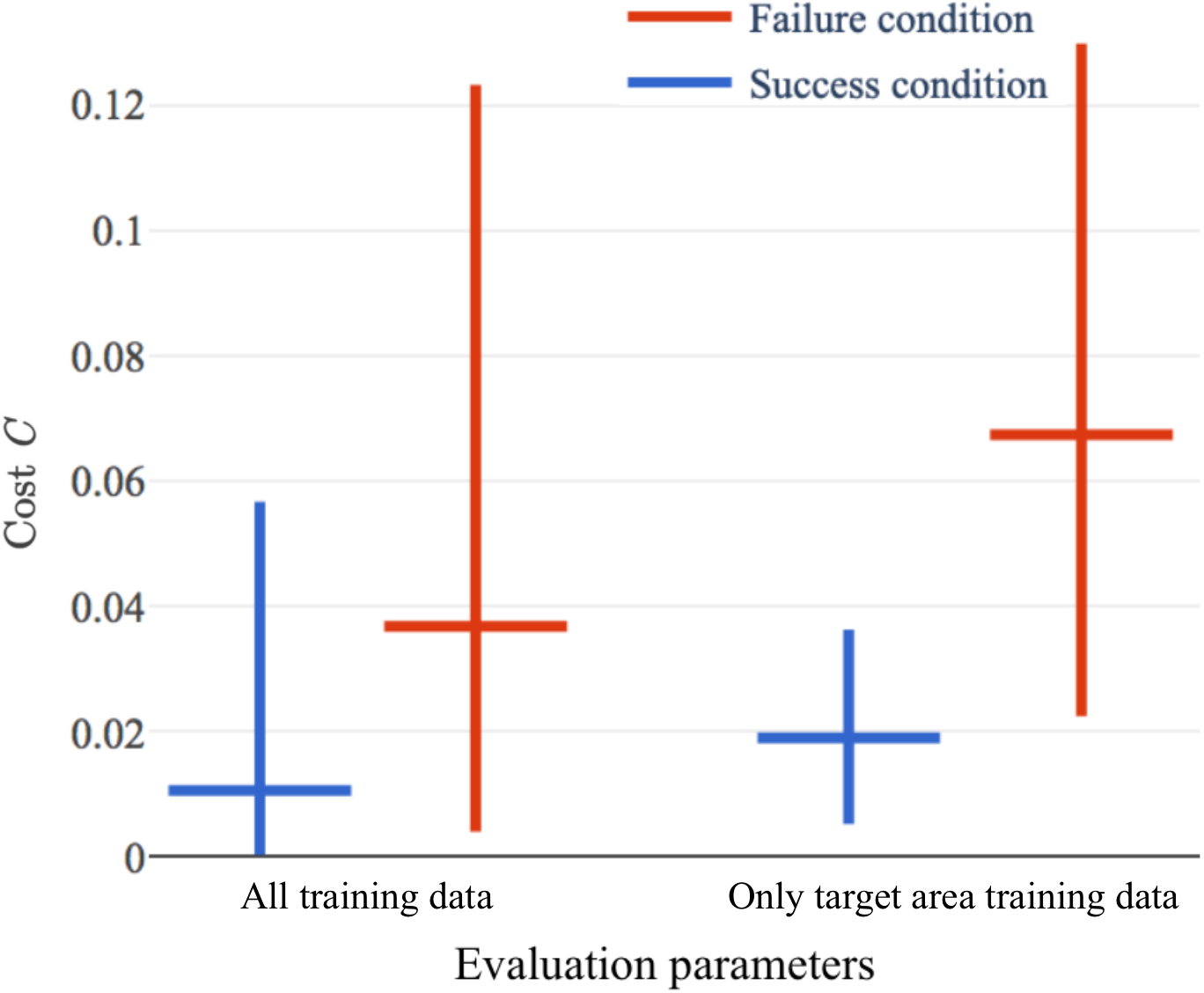}
    \caption{Results of the error evaluation for all test states grouped into trials. The horizontal lines indicate the mean value, corrected for different numbers of evaluable test states across single trials. The vertical lines indicate the average range along single trials in the condition.}
    \label{validationPlot}
\end{figure}

We then examined the same data grouped into trials, and the effect persists. The mean error values, when corrected for the different amounts of evaluable states in single trials, describe the same relation between the two conditions. The mean range along the trials also differed between the conditions just as expected from the overall comparison between the groups on both evaluations (Fig. \ref{validationPlot}). After a closer look at the descriptive data for each trial, we conclude that there is no reason to believe that the difference found in the overall comparison misrepresents the underlying data.

\begin{figure}[!t]
    \centering
    \includegraphics[width = 2.7in]{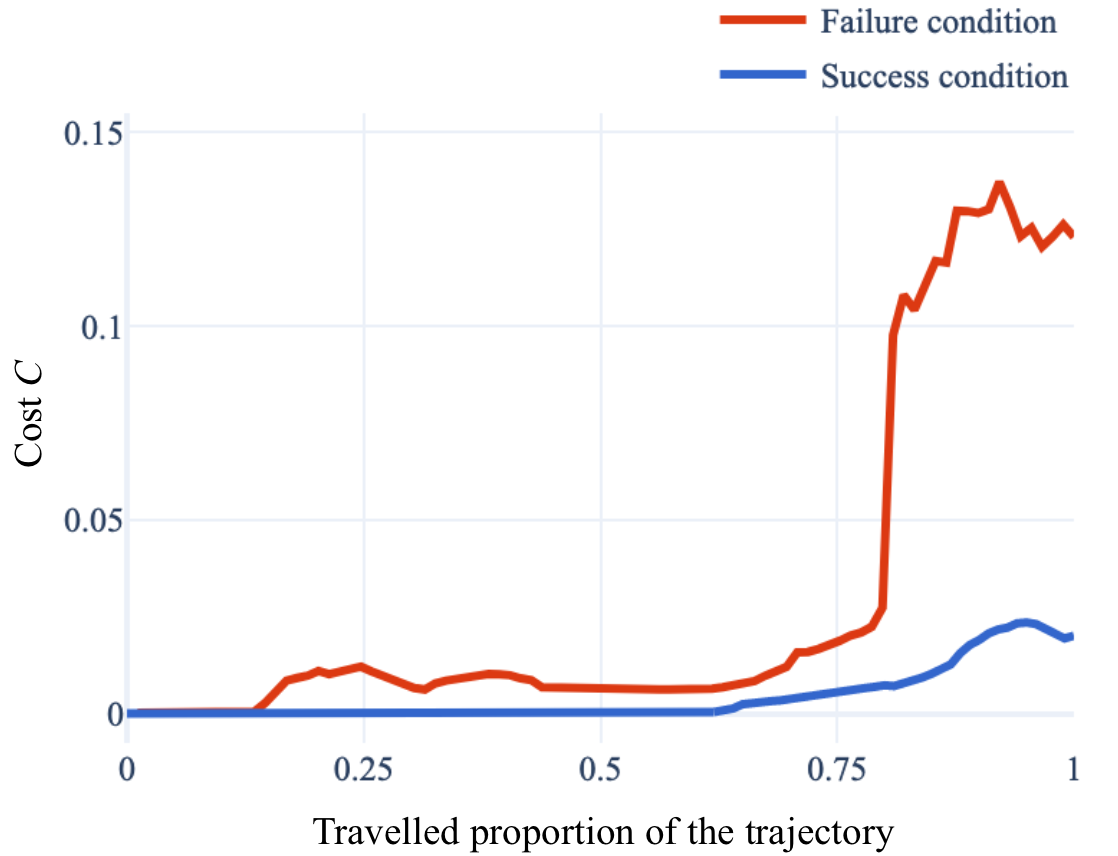}
    \caption{Error values along one single exemplary grasp trajectory of each experimental condition.}
    \label{fig:costAlongSingleTrials}
\end{figure}

The range in error values in the failure condition is substantially wider than in the success condition by a factor of 2.65 in the first evaluation, and 5.15 in the second evaluation. Fig. \ref{Boxplot} also shows that the difference between conditions is more pronounced in the second evaluation (based on only target area training data) than in the first evaluation (based on all training data).
Accordingly, we observed that the error development along single trials also shows a greater divergence between the two conditions towards the end of each reach-to-grasp trajectory (Fig. \ref{fig:costAlongSingleTrials}).

We also found that the maximal error value along each trial is reached very close to the object surface. In the first evaluation, across all grasp trials the highest error value is reached on average only 9.75 states away from the final grasp in the success condition, and 3.15 states in the failure condition. Similarly, in the second evaluation, the highest errors are on average 14.63 (success) and 4.85 (failure) states away from the grasp. All of these averages lie within the last 176ms or less of each trial.
Additionally, we found that the reported differences and especially the difference in range can be further tuned, by varying other parameters of the error function. As an example, the third group in Fig. \ref{Boxplot} displays error distributions for an evaluation using the target area training data and $n$ = 20.

Using a threshold of $C_T = 0.0249$ on our test data (evaluated on the target area training data with $n = 5$), the method correctly detected 18 out of 20 failures trials and correctly classified 6 out of 8 correct trials as non-failures (Sensitivity = 0.9, Specificity = 0.75).

\subsection{Based on rotational information only}
\begin{table}[!t]
  \centering
  \vspace{0.2cm}
  \caption{Sample sizes based on orientational information \newline of the wrist only.}
  \label{DataOverview2}
    \begin{tabular}{l c c}
    & \makecell[c]{Success\\condition} & \makecell[c]{Failure\\condition}\\ \hline
    Number of grasp trials &  8 & 20 \\
    \hline
    Total number of sampled test states & 1099 & 2066 \\
    \hline\makecell[l]{Number of evaluable test states based\\on the full demonstration data} & 1099 & 2066 \\
    \hline
    \makecell[l]{Number of evaluable test states based \\ on the target area demonstration data} & 1099 & 2051
    \end{tabular}
\end{table}

\begin{figure}[!t]
    \centering
    \vspace{0.2cm}
    \includegraphics[width = \linewidth]{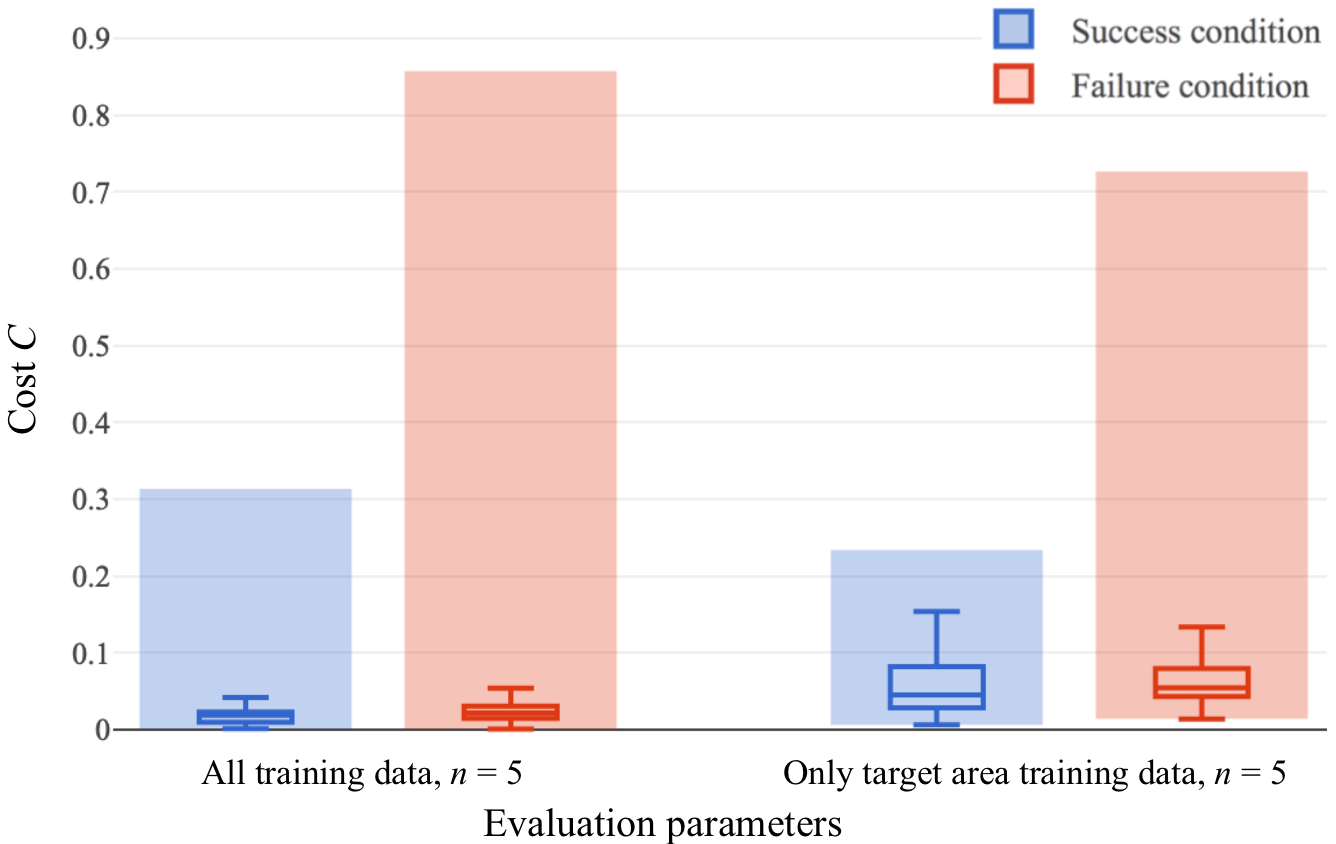}
    \caption{Comparison of error distributions between the two experimental conditions, based on only orientational information of the wrist. The chart shows Tukey boxplots, where the whiskers represent data within 1.5 inter-quartile-ranges of the first and third quartile. The bars illustrate the absolute range of error values.}
    \label{Boxplot2}
\end{figure}

In this second analysis the focal area of similar context was determined only by the angular distance between a test state and the training data. As a result, more training states were considered to be relevant to the current situational context. This greater generalisation increased the number of evaluable states (Table \ref{DataOverview2}), but at the same time slightly reduced the divergence between the two experimental conditions (Table \ref{ResultsTable2}, Fig. \ref{Boxplot2}). However, there is still a clear distinction between the success and the failure condition. Using a threshold value of $C_T =  0.0511$ on our test data (evaluated on all training data) the method achieved a sensitivity of 0.6, and a specificity of 0.875.

\begin{table}[!t]
    \centering
    \caption{Comparison of error distributions based on orientational information of the wrist only.}
    \label{ResultsTable2}
    \begin{tabular}{r c c}
    \textbf{Evaluation 1} & Success condition & Failure condition \\
    \hline
    $M$ &  0.0192 & 0.0436 \\
    $SD$ & 0.0231 & 0.1022 \\
    Median & 0.0189 & 0.0217 \\
    \hline
    $p$ & \multicolumn{2}{c}{$<$0.001} \\
    \end{tabular} \\
    \vspace{0.3cm}
    \begin{tabular}{r c c}
    \textbf{Evaluation 2} & Success condition & Failure condition \\
    \hline
    $M$ & 0.0533 & 0.0719 \\
    $SD$ & 0.0307 & 0.0802 \\
    Median & 0.0454 & 0.0546 \\
    \hline
    $p$ & \multicolumn{2}{c}{$<$ 0.001}
    \end{tabular} \\
    \vspace{0.3cm}
    \raggedright{\textit{Note.} Values reported in this table are the mean (\textit{M}), standard deviation (\textit{SD}), and the median of all calculated errors in each experimental condition, as well as the one-sided \textit{p}-value of a permutation test between the conditions after 5000 permutations.}
\end{table}

\section{Discussion}

The experimental results indicate that the proposed approach can potentially be used to monitor the status of a prosthetic control system, detect and even anticipate failures, and therefore increase its accuracy and reliability over time. The model can clearly distinguish between the two conditions; while lower values occur in both conditions, for high values it becomes more and more likely that the observed state is a failure, as there is a large range of error values that only ever occurred in the failure condition (see Fig. \ref{Boxplot}). The way to achieve an optimal failure detection must therefore be to minimise the overlap of the two conditions, or the area of uncertainty, and tune the error calculation in such a way, that the data in the two groups diverge as much as possible.


One simple way of determining whether we are facing a failure is then to use a threshold for triggering a model update; such a threshold will always constitute a trade-off between sensitivity and specificity, but the more distinctly we can isolate the two conditions, the more accurately and confidently we can place the threshold and identify failures. The exact value could even be tuned individually, depending on the available data and user preference. 

In this work we have made several assumptions. Firstly, the experiment was limited to two grasp types only, and secondly, the task only included a single object shape.
Additionally, the approach in this paper was tested only on one non-disabled expert user.
Future work will further explore the proposed method in a broader experiment with more participants, including patients from the target population. We also assumed that the user attempts a sensible grasping motion, as opposed to e.g. punching the object. In such cases the system will detect a failure, but the user should be able to dismiss the alert. 

As future extension we will also build on previous work of Kopicki and Zito \cite{A,stuber2018icra,zito2013iros,bib:zito_2019} who have demonstrated that a set of generative models can be efficiently learned, for a robot manipulator, in one shot such that manipulative contacts and trajectories are computed for previously unseen objects. By integrating these methods in our system, we will replace the human demonstration samples as a prior estimate of which grasps the user might attempt, and more importantly it would allow generalisation to new object shapes.

\section{Conclusion}

In this work we introduced an automatic failure detection method that continuously monitors the system performance. Although we are still in an early stage of development, our results indicate that the approach performs well and has potential for future extensions. We believe that this is a promising first step towards a truly reliable self-correcting control system, highly interactive and customisable, without the need for excessive training.
\bibliographystyle{IEEEtran}
\bibliography{MscReport}
\end{document}